\documentclass[aps,pre,twocolumn]{revtex4-1}
    \usepackage{amsmath}
    \usepackage{color}
    \usepackage{booktabs}  
    \usepackage{multirow}
    \usepackage{array}
    \usepackage{graphicx}
    \usepackage{diagbox}
    \usepackage{subfigure}
    \usepackage{caption}
    \usepackage{bbding}
\begin{document}
    \captionsetup[figure]{labelfont={bf},name={Figure},labelsep=period}
    \captionsetup[table]{labelfont={bf},name={Table},labelsep=period}
    \title{Highly Efficient Human Action Recognition with Quantum Genetic Algorithm Optimized SVM}
    \author{Yafeng Liu, Shimin Feng, Zhikai Zhao, Enjie Ding}
    
    \begin{abstract}
In this paper we propose the use of quantum genetic algorithm to optimize the support vector machine (SVM) for human action recognition. The Microsoft Kinect sensor can be used for skeleton tracking, which provides the joints' position data. However, how to extract the motion features for representing the dynamics of a human skeleton is still a challenge due to the complexity of human motion. We present a highly efficient features extraction method for action classification, that is, using the joint angles to represent a human skeleton and calculating the variance of each angle during an action time window. Using the proposed representation, we compared the human action classification accuracy of two approaches, including the optimized SVM based on quantum genetic algorithm and the conventional SVM with grid search. Experimental results on the MSR-12 dataset show that the conventional SVM achieved an accuracy of $ 93.85\% $. The proposed approach outperforms the conventional method with an accuracy of $ 96.15\% $.
    \end{abstract}

    \maketitle
    
    \section{Introduction}
    Human action recognition (HAR) plays an important role in video surveillance, health care and human computer interaction(HCI) \citep{chernbumroong2014practical}. One goal of HAR is to provide the information about the user's actions with the help of a computer. The information can be widely applied in artificial intelligence area. For example, the recognition and prediction of elderly people's actions will help them with their health care \citep{abowd1998context}. Human activity recognition also plays a key role in natural interaction area in HCI. Moreover, HAR bring a new vision to some traditional areas, e.g. sports motion analysis, virtual reality (VR), augmented reality (AR) and other human-computer interaction area.
    
     According to the classical Newtonian mechanics theory, people can get the kinematic law of the object completely when they know the initial state and driving force. However, this method does not work here, since the actions of human body is complex and furthermore there is a large number of interactions. So the actions pattern is not easy to be described \citep{aggarwal2011human}, we must ask some tools for help.  The common research tools can be divided into two categories, including the video-based methods and the sensor-based methods \citep{woznowski2016classification}. We will use  sensor-based one here. The development of new sensing devices, e.g. the Microsoft Kinect and other RGB-D devices bring new opportunities for the HAR researchers \citep{presti20163d}. This kind of devices are inexpensive, portable, and can be used for skeleton tracking, which provide $ 15-20 $ joints' information. One question we should mention that sample inputs will be considered since some good sample representation can make problem simple and accuracy.  Joint positions, key poses and joint angles is some usual sample representations. This paper presents an approach for features extraction that considers only the information obtained from the 3-dimensional skeletal joints. We extract the skeletal features by computing all angles between any triplet of joints and then calculate the variance of each angle during the time period when an action is performed.
     
     Another question is that how to choose an effective pattern recognition algorithm. people have tried many methods, such as Decision Tree(DT), Bayes methods, k-Nearest Neighbour(kNN), Neural Network(NN), Support Vector Machine(SVM), Hidden Markov Model(HMM) and so on\citep{seddik2017human,gaglio2015human}. Among them Support vector machines are widely used because of their simplicity and efficiency. Support Vector Machine \citep{Cortes1995Support} classifies the data by constructing hyperplane, separating different categories of data from each other. Nevertheless, it is not an easy task to find the appropriate parameters for SVM due to the limited searching capability with the grid search method. Thus the best classification results can not be achieved. An inappropriate parameter will decrease the performance of SVM classifier, so people have tried some methods for optimized parameters. Grid search, particle swarm algorithm and genetic algorithm are common used here.

    Here we will use the quantum genetic algorithm to improve the efficiency of SVM parameter optimization. The quantum algorithm is based on the correlation \citep{mosca2008quantum,nielson2000quantum,jones2013computing} of quantum bits, which gives the algorithm the characteristics of parallelism. Compared with the classical algorithm, the computational efficiency has been greatly improved \citep{lenstra2000integer,jones2013computing}. Since the improvement of the search efficiency, the population search range of SVM parameters is enlarged. In recent years, the quantum genetic algorithms have been widely used in machine fault diagnosis, geology research and environmental analysis \citep{zhang2017screw,wei2016fault,chen2016an,zhou2010forecasting, Xie2015stability}. In this paper, we use the quantum genetic algorithm to optimize the SVM for classifying the human actions by  building a better SVM model.
    
    The rest of the paper is organized as follows. Section $ 2 $ presents the Kinect system and the classification algorithm. We describe the experimental results in Section $ 3 $ and concludes the paper in Section $ 4 $.
    
    \section{Methods}
\subsection{Feature Extraction of Human Action}
 
    \begin{figure*}[htbp!]
    	\centering
    	\subfigure[The skeleton joints]{
    		\begin{minipage}[b]{0.45\textwidth}
    			\includegraphics[width=1\textwidth]{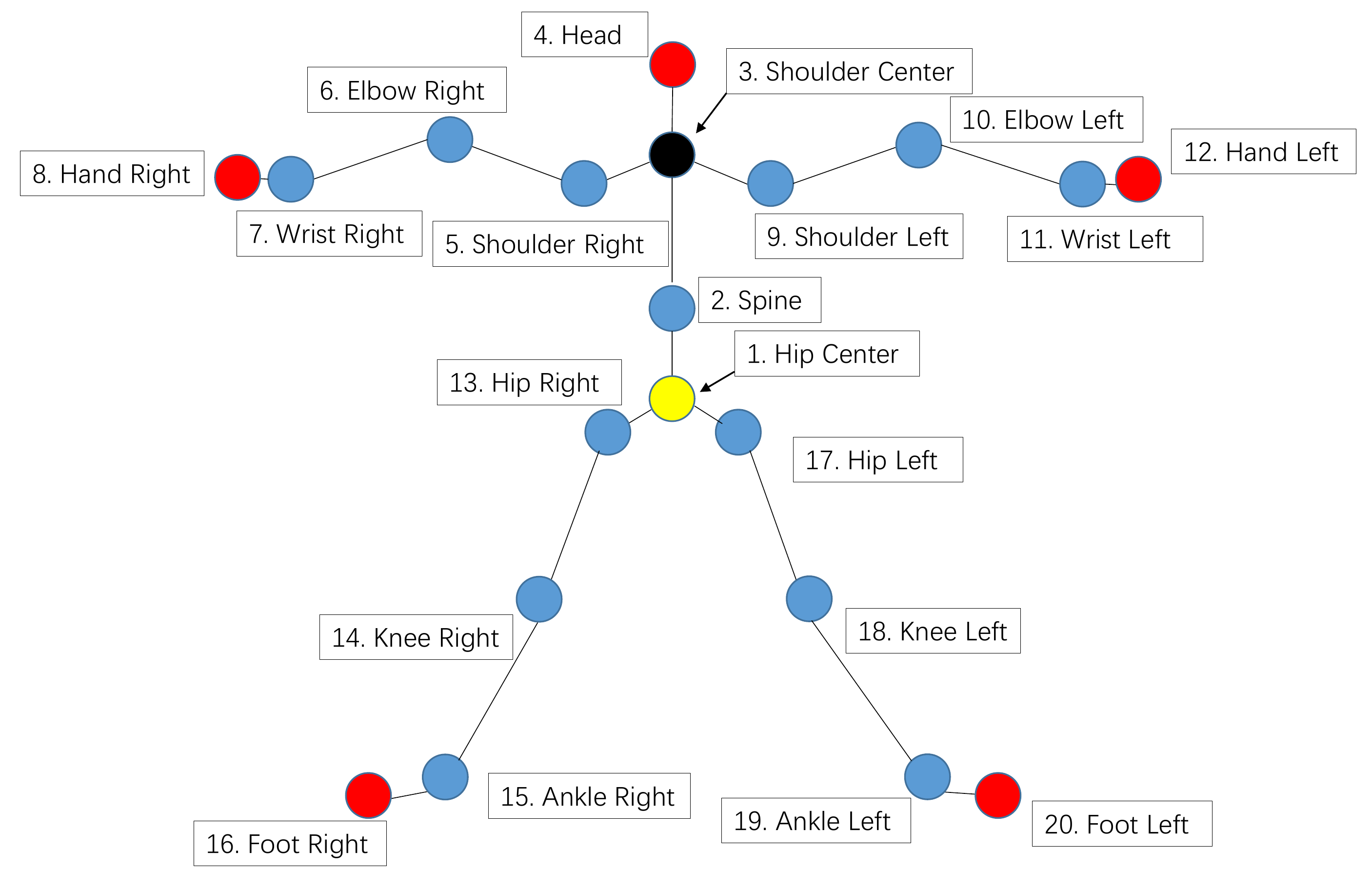}
    			\vspace{-2em}
    			\label{full_system}
    		\end{minipage}
    	}
    	\subfigure[Special label of angel sample within the part of the hip and the shoulder center.]
    	{
    		\begin{minipage}[b]{0.45\textwidth}
    			\includegraphics[width=1\textwidth]{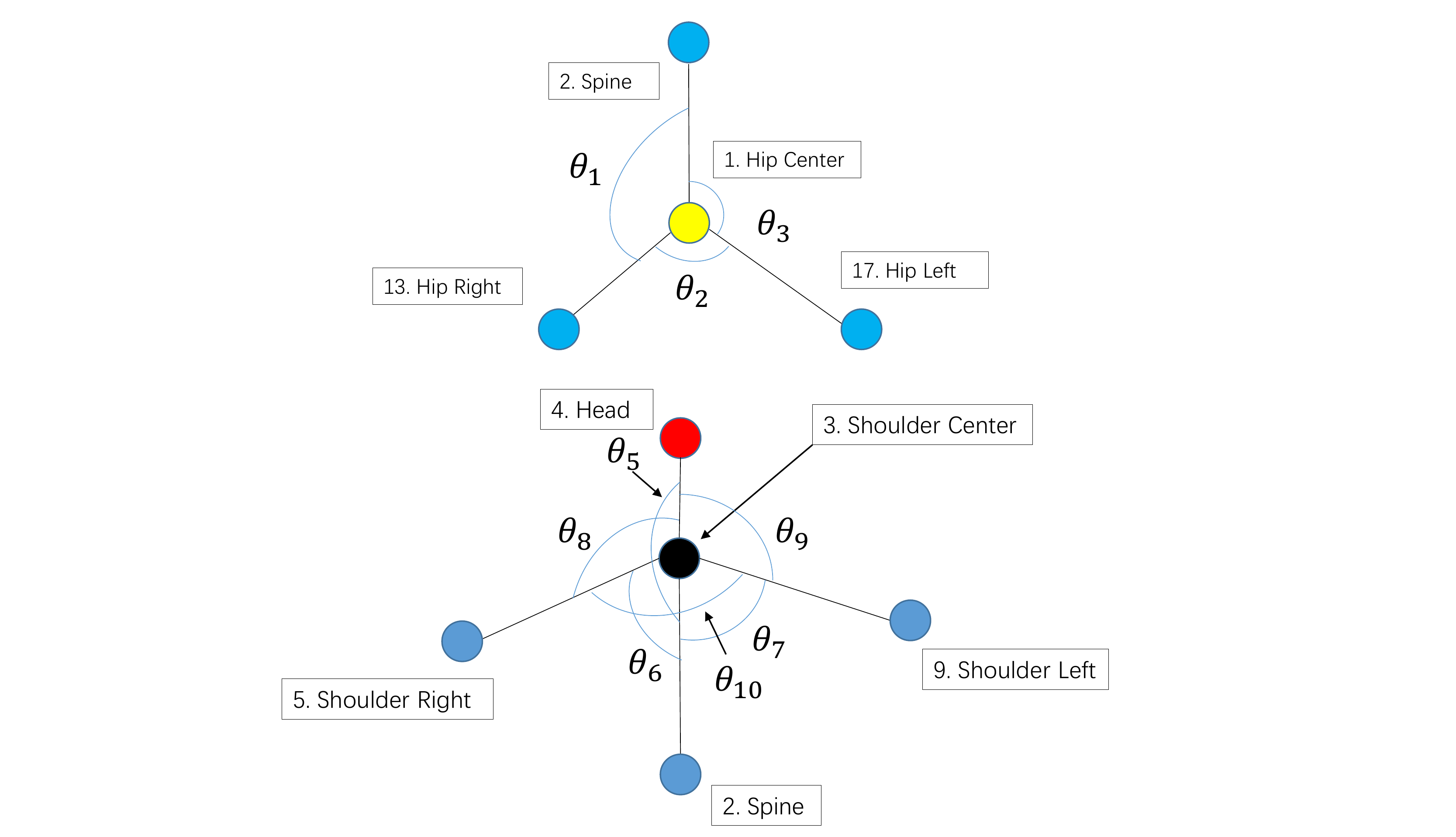}
    			\vspace{-2em}
    			\label{hip}
    		\end{minipage}
    	}
    	\caption{The schematic diagram of the skeleton joints} \label{joint}
    \end{figure*}
    
    As shown in Figure~\ref{full_system}, we get the 3D positions of each skeleton joint with the Kinect. The human action sample is defined as
    
    \begin{equation}
    \boldsymbol{f}=[\boldsymbol{f_1},\boldsymbol{f_2},\dots,\boldsymbol{f_n},\dots,\boldsymbol{f_{20}}]
    \end{equation}     
    
    where $\boldsymbol{f_n}$ is the coordinates of the $ n $-th joint. As there are $ 20 $ joints, we get a 60-dimensional vector. In reality, we don't need an accurate joint position for human activity recognition as the relative positions will meet our requirements. We calculate the relative position of the two adjacent joints. This vector illustrates the direction of the limb between these two joints. For a joint connected with multiple limbs, the angles are formed, as shown in Figure~\ref{hip}.

    Consider a skeleton joint and the two adjacent joints, the coordinates are
    \begin{subequations}\label{position}
    	\begin{align}
    	\boldsymbol{f_{n-1}} &= (x_{n-1},y_{n-1},z_{n-1})\\\label{position:sub1}
    	\boldsymbol{f_n} &= (x_n,y_n,z_n)\\
    	\boldsymbol{f_{n+1}} &= (x_{n+1},y_{n+1},z_{n+1}) \label{position:sub3}
    	\end{align}  
    \end{subequations}

    The vector of the limb is defined by two adjacent joints:
    
    \begin{subequations}
    	\label{limb}
    	\begin{align}    
    	\boldsymbol{a} &= \boldsymbol{f_{n-1}}-\boldsymbol{f_{n}}\\
    	\boldsymbol{b} &= \boldsymbol{f_{n+1}}-\boldsymbol{f_{n}} \label{limb:sub2}
    	\end{align}  
    \end{subequations}
    
    The angle $\theta$ formed by these two limb is
    \begin{equation}
    \theta=\frac{\cos(\boldsymbol{a},\boldsymbol{b})}{||a||*||b||}
    \end{equation}
    
     As Fig~\ref{full_system} shows, there are five joints which own only one junction (red points in Fig~\ref{full_system}), and no angles exist on these joints. There are 13 joints which own two junctions (blue points in Fig~\ref{full_system}), and each of these joints owns one angle. There are only one joint which owns three junctions (yellow point in Fig~\ref{full_system}). According to the knowledge of permutation and combination, the angles on this joint are $C_3^2=3$ totally. At last, the shoulder center joint own four junctions(black point in Fig~\ref{full_system}). So there are $C_4^2=6$ angles on it. The total number of angles is $13+3+6=22$ on body joints. 
     
     The relationship between the skeleton joints and the angles is shown in Table~\ref{mapping}. The left column in the table shows the spatial position label of joints, and they are all three dimensional vector. The right column shows the the intersection angle label between limb, and all of them are scalars. These labels of angle sample are arranged according to their order of position sample. We can see that the dimensions of sample are reduced from 60 to 22 using angle strategy. However, there are two special joint, joint 1 and joint 3(see Fig~\ref{hip}). More than one angles exist within these two joints. We define another ranking method here: we arrange angles within the same joints according to the order of joint label next to it. For example, the first joint, hip center, there are three connecting joints, which are the 2nd, the 13th and the 17th joint, as in the upper part of the Fig~\ref{hip}: We name the angle formed by joint 2 and joint 13 $\theta_1$, the angle formed by joint 2 and joint 17 $\theta_2$, and the joint formed by joint 13 and 17 $\theta_3$. For the joint 3, we use the same method dealing with it(see the lower part of fig~\ref{hip}). 
     
    \begin{table}[ht!] 
    	\centering   
    	\caption{The corresponding relationship between the skeleton joints and the angles. From the beginning of the forth joint, the joint connected to only one joint will exist every other three joints periodically, and it can't form any angles. All the other joints connect to two joints, and form one angle. The table doesn't show all mapping relations, and the omitted part is represented by the ellipsis.} 
    	\label{mapping}  
    	\begin{tabular}{|c|c|}\hline   
    		The index of skeleton joints & The index of joint angles \\\hline
    		1&$\theta_1, \theta_2, \theta_3$\\\hline
    		2&$\theta_4$\\\hline
    		3&$\theta_5, \theta_6, \theta_7, \theta_8, \theta_9, \theta_{10}$\\\hline
    		4&---\\\hline
    		5&$\theta_{11}$\\\hline
      		\multicolumn{2}{|c|}{$\vdots$}\\\hline
    		19&$\theta_{22}$\\\hline
    		20&---\\\hline
    	\end{tabular} 
    \end{table} 
    
    We use the angle representation method to reduce the $ 60 $-dimensional vector to $ 22 $-dimensional vector. For the continuous action, we need to add the timing information and process multiple frames together for action recognition. Here we set a time window for action segmentation. Assume the action lasts for $ T $s, during which period there are $ M $ frames acquired from the Kinect. We can get the variance of each angle in this time window.
    
    \begin{equation}
    \boldsymbol{j_n}=\frac{1}{(M-1)}\sum_M^k(f_{nk}-\mu_n)^2
    \end{equation}
    
    \subsection{Quantum Genetic Algorithm (QGA)}
    Quantum genetic algorithm is an optimization algorithm based on quantum computing theory. The basic representation in the quantum theory is a coherent state, which is very different from classical one. Here, we use the state vector to describe genetic coding, and use the quantum logic gate to realize the evolution of population. Because of the kind of representation, quantum algorithm has the characteristics of parallelism, and for this reason it is more faster than the traditional algorithm in p searching speed.
    
    \textbf{The Coding of Quantum genetic algorithm}. The binary and decimal codes are used in the classical genetic algorithm. When quantum bits are used, the encoding will be different. There is superposition and coherence between the quantum states, so unlike the classical bits, there are entanglement properties in the quantum bits. For a quantum bit, it cannot simply be written as 0 or 1 states, but as an arbitrary superposition between them, so the quantum bit can be written as:
    
    \begin{equation}
    \vert\psi\rangle = \alpha\vert 0\rangle+\beta\vert 1\rangle
    \label{qubit}
    \end{equation}
    
    where $|0>$ and $|1>$ are both vectors, representing the system states. $\alpha$ and $\beta$ are a pair of parameters, and the square of them corresponds to the probability measuring of these two states. These two parameters satisfy the normalization rule:
    
    \begin{equation}
    \left| \alpha_i \right|^2 + \left| \beta_i \right|^2 = 1
    \label{nomalization}
    \end{equation}
    
    A chromosome with $m$ bits can be expressed as Eq.~\ref{chromosome}, and for each element of the chromosome, $|\alpha_i|^2 + |\beta_i|^2 = 1, i=1, 2, \cdots, m$。
    
    \begin{equation}
    P_j = \left[  \begin{array}{cccc}
    \alpha_1 & \alpha_2 & \cdots & \alpha_m \\
    \beta_1 & \beta_2 & \cdots & \beta_m \end{array} \right]
    \label{chromosome}
    \end{equation}
    
    \textbf{The Quantum logic gate}. In the quantum genetic algorithm, the operation of quantum bits is achieved through the quantum logic gate. The quantum logic gate can help realize the evolution of the population. The optimal gene can be produced through the guidance of rotation strategy(see Table~\ref{rotation_strategy}). This can speed up the entire algorithm. The operation of a quantum logic gate can be expressed in the form of a matrix:
    
    \begin{equation}
    \left[ \begin{array}{c}
    \alpha_i^{t+1} \\
    \beta_i^{t+1} \end{array} \right] = G\left[ \begin{array}{c}
    \alpha_i^t \\
    \beta_i^t \end{array} \right]
    \end{equation}
    
    where $[\alpha_i^t,\beta_i^t]$ and $[\alpha_i^{t+1},\beta_i^{t+1}]$ represent the quantum bits of the chromosomes for the generation $t$ and $t+1$ respectively. $ G $ represents the quantum logic gate:
    \begin{equation}
    G = \left[ \begin{array}{cc}
    \cos\theta_i & -\sin\theta_i\\
    \sin\theta_i & \cos\theta_i     
    \end{array}
    \right]
    \end{equation}
    
    $\theta$ is the rotation angle. The selection of direction and magnitude is shown in Table~$\ref{rotation_strategy}$.
    
    \begin{table*}[htbp] 
    	\centering   
    	\caption{The rotation strategy of the quantum logic gate.} 
    	\label{rotation_strategy}  
    	\begin{tabular}{p{1cm}p{1cm}<{\centering}p{3cm}<{\centering}p{1.5cm}p{2cm}<{\centering}p{2cm}<{\centering}p{2cm}<{\centering}p{2cm}<{\centering}}  
    		\toprule  
    		\multirow{2}{*}{$\boldsymbol{x_j}$}&\multirow{2}{*}{$\boldsymbol{best_j}$}&\multirow{2}{*}{$\boldsymbol{f(x)>f(best)}$}&\multirow{2}{*}{$\boldsymbol{\Delta\theta_j}$}&  
    		\multicolumn{4}{c}{$\boldsymbol{s(\alpha_j,\beta_j)}$}\cr 
    		\cmidrule(lr){5-8}
    		&&&&$\boldsymbol{\alpha_j\beta_j>0}$&$\boldsymbol{\alpha_j\beta_j<0}$&$\boldsymbol{\alpha_j=0}$&$\boldsymbol{\beta_j=0}$\cr  
    		\midrule  
    		\textbf{0}&\textbf{0}&\textbf{FALSE}&\textbf{0}&\textbf{0}&\textbf{0}&\textbf{0}&\textbf{0}\cr  
    		\textbf{0}&\textbf{0}&\textbf{TRUE}&\textbf{0}&\textbf{0}&\textbf{0}&\textbf{0}&\textbf{0}\cr  
    		\textbf{0}&\textbf{1}&\textbf{FALSE}&$\boldsymbol{0.01\pi}$&\textbf{+1}&\textbf{-1}&\textbf{0}&$\boldsymbol{\pm1}$\cr  
    		\textbf{0}&\textbf{1}&\textbf{TRUE}&$\boldsymbol{0.01\pi}$&\textbf{-1}&\textbf{+1}&$\boldsymbol{\pm1}$&\textbf{0}\cr  
    		\textbf{1}&\textbf{0}&\textbf{FALSE}&$\boldsymbol{0.01\pi}$&\textbf{-1}&\textbf{+1}&$\boldsymbol{\pm1}$&\textbf{0}\cr  
    		\textbf{1}&\textbf{0}&\textbf{TRUE}&$\boldsymbol{0.01\pi}$&\textbf{+1}&\textbf{-1}&\textbf{0}&$\boldsymbol{\pm1}$\cr  
    		\textbf{1}&\textbf{1}&\textbf{FALSE}&\textbf{0}&\textbf{0}&\textbf{0}&\textbf{0}&\textbf{0}\cr
    		\textbf{1}&\textbf{1}&\textbf{TRUE}&\textbf{0}&\textbf{0}&\textbf{0}&\textbf{0}&\textbf{0}\cr 
    		\bottomrule  
    	\end{tabular} 
    \end{table*}  
    
    In Table~$\ref{rotation_strategy}$, $x_i$ and $b_i$ represent the optimal chromosome and the current optimal chromosome, respectively. $f(x) $ is the fitness function, $\delta\theta$ is the rotation angle. By selecting different rotation angles, we can control the convergence speed and accurate.
    
    \subsection{Support Vector Machine}
    \begin{figure}[htbp]
    \centering
    \includegraphics[width=0.4\textwidth]{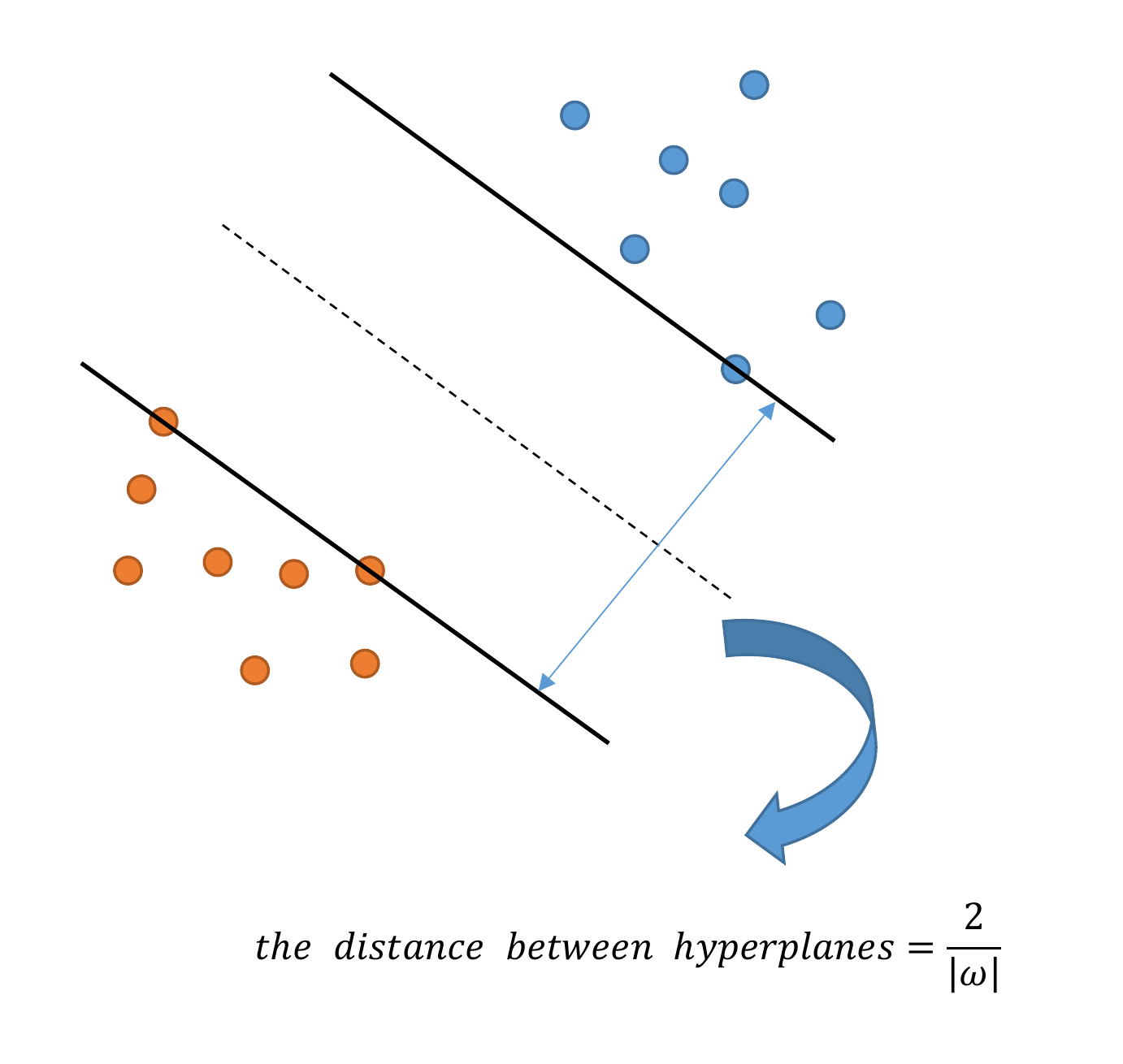}
    \caption{The diagram of the classification surface}
    \label{SVM_model}
    \end{figure}
    
    Support Vector Machine (SVM) can mainly be divided into two parts: classification and regression.  It is widely used for its highly efficiency and simplicity. For classification problem, it is a kind of supervised learning model which classifies the representation $x_k \in R^n$ of an object in high dimensional space according to a label $y_k \in R$. This representation and label constitute the sample space $d={(X_k, y_k) |k = 1, 2, \cdots, N}$, where $N $ is the number of the samples.Support Vector machine is to find a pair of hyperplanes, which  separately passes through the nearest two points in different classes. In order to achieve the best classification results, we need to make the distance between the hyperplanes as large as possible. As shown in Figure~\ref{SVM_model}, the hyperplanes are represented by the solid line. Therefore, the task of SVM is simply boiled down to find the maximum value of $2/|\omega|$ in this figure. This optimization problem can be written as:
    
    \begin{equation}
    \min\limits_{\boldsymbol{\omega},b} \boldsymbol{\psi}(\boldsymbol{\omega},b) = \frac{1}{2}\boldsymbol{\omega}^T\boldsymbol{\omega}
    \label{taget_function}
    \end{equation}
    
    Where $\boldsymbol{\omega}$ and $b$ represent the slope and intercept of hyperplanes. Considering the constraint of hyperplanes passing through the closest points $y_k (\boldsymbol{\omega}^t\phi (x_k)+b) \geq 1$, the Lagrange equation can be obtained:
    
    \begin{equation}
    L(\boldsymbol{\omega},b,\boldsymbol{\lambda})= \boldsymbol{\psi}(\boldsymbol{\omega},b) - \sum_{k=1}^N \lambda_k \{y_k(\boldsymbol{\omega}^T\phi(x_k)+b)-1 \}
    \label{Lagrange_equation_1}
    \end{equation}
    
    $\boldsymbol{\lambda}$ here is Lagrange operator. The parameters of SVM can be obtained by finding the extreme values of the equation. However, for practical problems, the distance between two classes may not be so large. Thus, it is necessary to introduce the concept of soft interval classification, that is, to allow some points to fall between two of hyperplane, but not across the middle dotted line. In this case, the target function needs to add a slack variable and the constraint should be modified to $y_k (\boldsymbol{\omega}^t\phi (x_k) +b) \geq 1-\epsilon_k$, $\epsilon_k \geq 0$. $\epsilon_k$ is the kth relaxation variable. This condition is not so strict as the previous constraint. The sample point may appear between the two hyperplanes. The corresponding dual equation can be modified to:
    
    \begin{multline}
    L(\boldsymbol{\omega},b,\boldsymbol{\epsilon},\boldsymbol{\lambda},\boldsymbol{\mu}) = \boldsymbol{\psi}(\boldsymbol{\omega},b) + C \sum_{k=1}^N \epsilon_k^n \\
    - \sum_{k=1}^N \lambda_k \{y_k(\boldsymbol{\omega}^T\phi(x_k)+b)-1+\epsilon_k \} - \sum_{k=1}^N \mu_k\epsilon_k
    \label{Lagrange_equation_2}
    \end{multline}
    
    In the Eq~\ref{Lagrange_equation_2}, $c$ is the penalty factor and $n $ is a natural number, corresponding to the $n$-order soft interval classification. $\boldsymbol{\lambda}$ and $\boldsymbol{\mu}$ here are Lagrange operator. We set $n=1$ here, which is the linear soft interval classification. By some commonly used derivation methods, we can get a simplified equation:
    
%
%
%
    
    \begin{equation}
    L(\boldsymbol{\lambda})=\max\limits_{\boldsymbol{\lambda}} \sum_{k=1}^N \lambda_k - \frac{1}{2}\sum_{i=1}^N\sum_{j=1}^N\lambda_i\lambda_j y_i y_j \phi(x_i) \phi(x_j)
    \label{simplify_lagrange}
    \end{equation}
    
    In comparison with the conventional SVM, the constraint conditions are changed:
    
    \begin{align}
    &\sum_{k=1}^N \lambda_k y_k= 0 \\
    &0 \leq \lambda_k \leq C \quad k=1,2,\cdots,N    
    \end{align}
    
    When describing the parameters of the sample points, we do not use $x_k$ directly, instead we use a mapping $\phi (x_k) $. This is due to the fact that we cannot get good classification results with a linear classification plane in many practical problems. We need a more complex plane to make the classification better. This kind of mapping plays such a role. In $\ref{simplify_lagrange}$, we define $K (x_i,x_j) =\phi (x_i) \phi (x_j) $, where $K (x_i,x_j) $ is called the kernel function. The commonly used kernel functions are listed follow:
    
    \paragraph{Radial Basis Function} 
    \begin{equation}
    K(x_i,x_j)=exp(\Vert x_i-x_j\Vert^2/\boldsymbol{\sigma}^2)
    \end{equation}
    
    \paragraph{Polynomial Kernel Function} 
    \begin{equation}
    K(x_i,x_j)=(x_i\cdot x_j+c)^d
    \end{equation}
    
    \paragraph{Sigmoid Kernel Function}
    \begin{equation}
    K(x_i,x_j)=\tan(k(x_i\cdot x_j)+v)
    \end{equation}
    
    \paragraph{Linear Kernel Function}
    \begin{equation}\label{core_function}
    K(x_i,x_j)=x_i\cdot x_j
    \end{equation}      
    
    \subsection{The Flowchart of the Algorithm}  
    
\begin{figure}[!t]
    	\centering
    	\includegraphics[scale=.5]{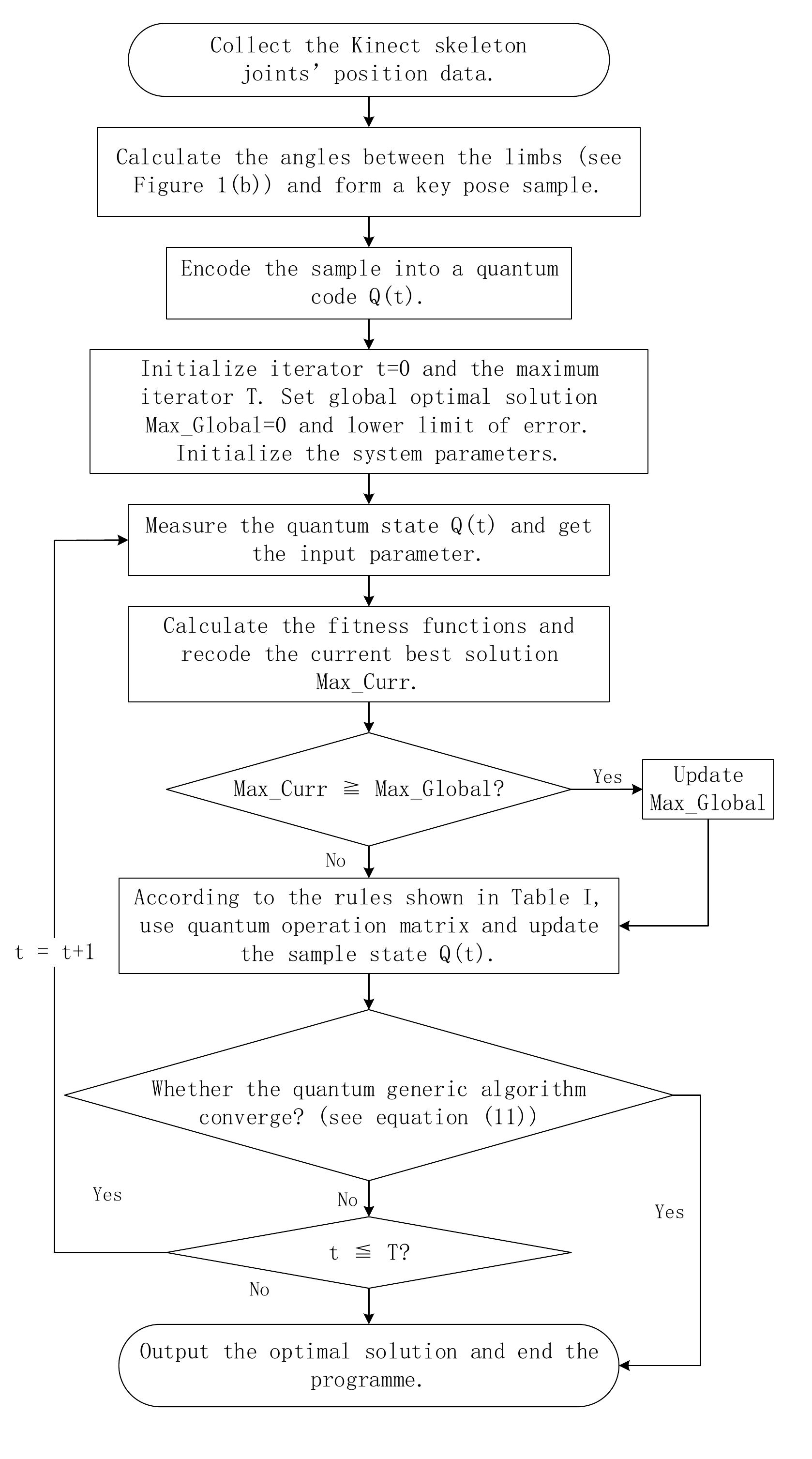}
    	\caption{Flowchart}
    	\label{flow_chart}
    \end{figure}

    We will use radial basis function for next research. In order to make the support vector machine run normally, the penalty factor $c$ and the kernel function parameter $\sigma$ are two variables needed to be determined according to $\ref{simplify_lagrange}\sim\ref{core_function}$. These two variables will directly affect the accuracy of the classification. How to determine these two parameters quickly and accurately is the key to the successful SVM model. Therefore, we will use the more efficient quantum genetic algorithm to help find these two parameters. The flowchart is shown in Figure~\ref{flow_chart}. It can be seen that the two parameters need to be quantum encoded first. Then the optimal solution is constantly adjusted through the quantum logical gate. By initializing a set of system parameters, we can calculate the classification accuracy. This accuracy can be used as the fitness function. We aim to search out a set of $(C,\sigma)$ according this fitness function.
    
    \begin{figure*}[htbp!]
    \centering
    \includegraphics[width=0.7\textwidth]{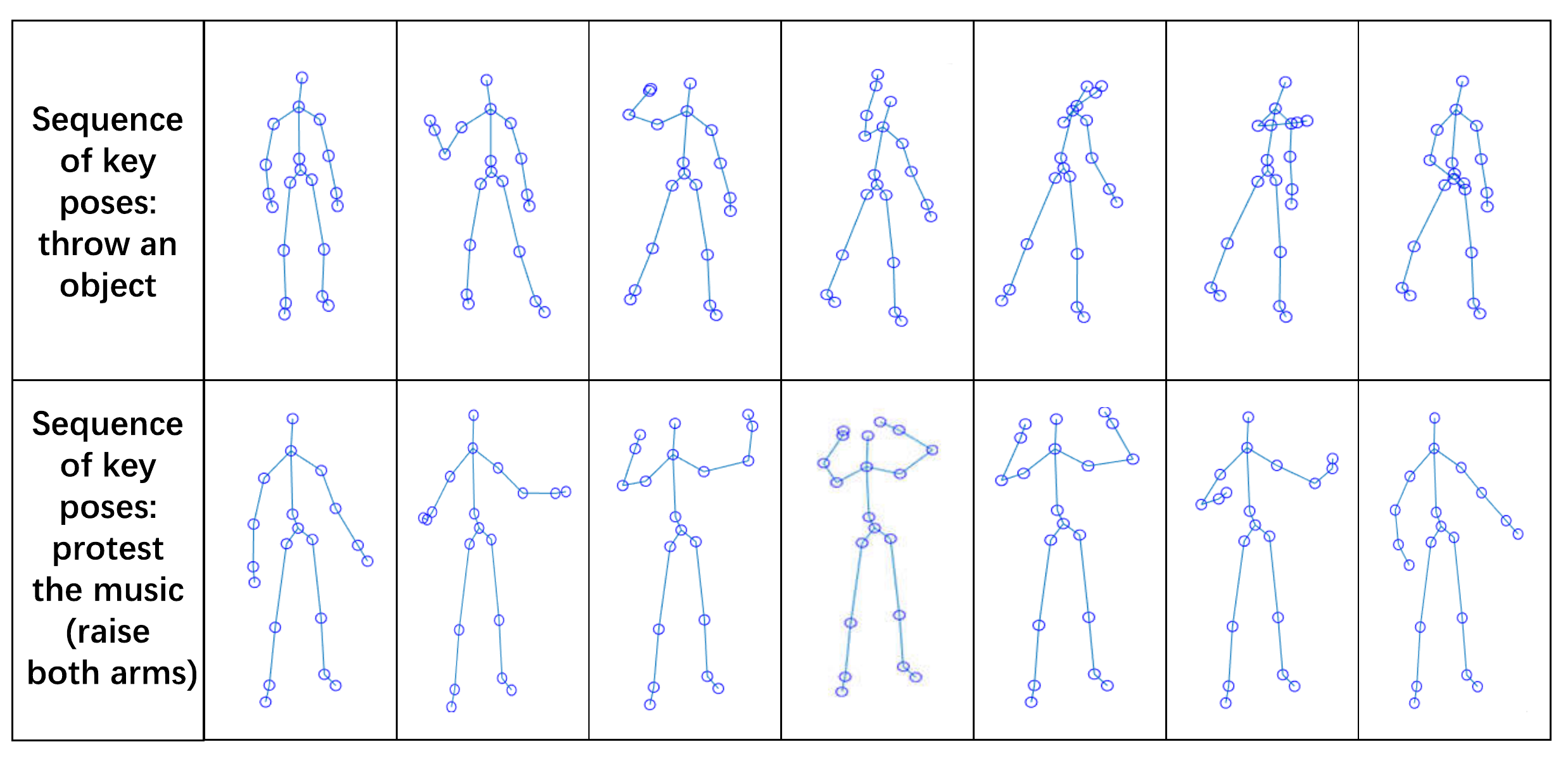}
    \caption{Action Segmentation: we use the MSRC-12 dataset collected by the Cambridge Microsoft Lab. We segment a complete action from the whole , the upper panel shows a segmentation of the "throw" action. The lower panel shows the "raising both arms" action.}
    \label{individual_component}
    \end{figure*}
    The eight steps of the SVM based on QGA optimization algorithm:
    
   \textbf{Step 1} Initialize the algorithm parameters, including the maximum number of iterations, population size, variable binary length and so on. Enter the training set data and test set data, as well as the corresponding labels.
   
   \textbf{Step 2} Initialize population $q (t) $ of penalty factor and parameters of kernel function: equal treatment of all genes, that is, initialize all genes $[\alpha_i^t,\beta_i^t]$ to $[1/\sqrt{2},1/\sqrt{2}]$, indicating that each chromosome appears equally in the initial search.
   
   \textbf{Step 3} Measure the initial population and get a specific $ p (t) $, which is a series of binary codes of the initialization length. Change them into decimal number and bring them into the SVM model with the training sample. The current individual is evaluated and the optimal individual is retained.
   
   \textbf{Step 4} Determine if precision is convergent or if the maximum number of iterations is reached. If yes, the algorithm terminates, else go to step 5.
   
   \textbf{Step 5} Update population $q (t) $ by using the rotation angle strategy in table $\ref{rotation_strategy}$.
   
   \textbf{Step 6} Check to see whether the catastrophic conditions are met. If yes, keep the optimal value and re-initialize the population. If not, go to step 7.
   
   \textbf{Step 7} Increase the number of iterations by one and return to step 3 to continue the execution.
   
   \textbf{Step 8} Output the optimization parameters and evaluate the test samples with these parameters.
    
    \section{The Classification Process and Experiment Results}
    
    \subsection{Problem Statement}
    \begin{figure}[t!]
    \centering
    \includegraphics[width=0.5\textwidth]{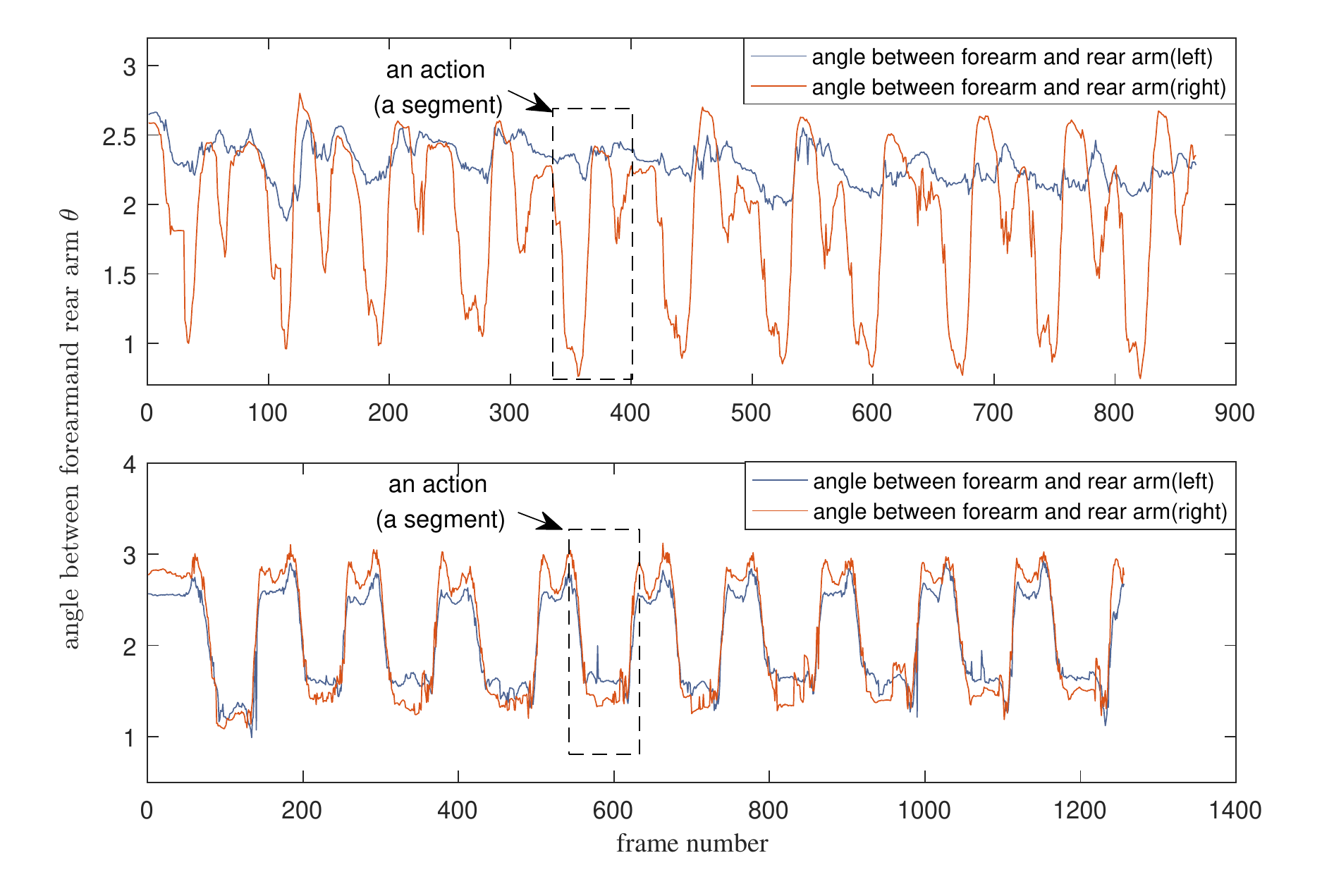}
    \caption{The curve of the elbow angle changes. It is the angle between the forearm and the back arm of the elbow. The picture above is the action "throw", the lower figure shows the raising both arms movement. The motion amplitude of the left arm of the throwing action is much smaller than that of other limbs.}
    \label{action_fragment}
    \end{figure}    
    We used the MSRC-12 gesture dataset \cite{fothergill2012instructing}, which consists of sequences of 12 groups of actions collected by the Cambridge Microsoft Laboratory through the $kinect$ system. We selected the eighth group of holding the hand and the ninth group of protest the music two similar movements to carry on our research. The segmentation of both actions is shown in Figure \ref{individual_component}. As mentioned above, the Kinect collection method is to record the three-dimensional real-time coordinates of the human joints as shown in Figure $\ref{full_system}$. Further, we calculated the angle of the torso on each joint point through the these data. The $1\sim9$ represents the relative angle between the limbs of torso, and $10\sim16$ indicates the relative angle between the limbs of the upper body, and the relative angle between the limbs of the lower body is $17\sim22$.

    Figure \ref{action_fragment} shows the change of the angle of limbs at the elbow in both arms. As can be seen from the figure, the curve periodically renders 10 sets of actions. We can see that the magnitude of the left arm movement changes during the throwing motion, which is much smaller than the other three curves. Therefore, the angle changing amplitude of the two arms can be used as a basis to distinguish the two kinds of motion patterns. 
    
    \subsection{Results}
    \renewcommand{\multirowsetup}{\centering}
    \begin{table*}[htbp!] 
	\centering   
	\caption{The parameters obtained through the cross validation (CV) and the quantum genetic algorithm (QGA), respectively.} 
	\label{result}  
	\begin{tabular}{|c|c|c|c|c|c|}\hline   
		&penalty&kernel function &\multirow{1}[4]{*}{generation}&\multirow{1}[4]{*}{accuracy}&\multirow{1}[4]{*}{time$(S)$}\\
		&factor C&parameter $\sigma$&&&\\\hline
		\multirow{2}{1.5cm}{cross validation}
		&\multirow{2}*{0.25}&\multirow{2}*{0.0625}&\multirow{2}*{\XSolid}&\multirow{2}*{93.85$\%$}&\multirow{2}*{4.38}\\
		&&&&&\\\hline
		\multirow{4}{1.5cm}{quantum \\ genetic algorithm}
		&15.839&0.155&1&93.85$\%$&6.83\\\cline{2-6}
		&10.831&0.080&2&96.15$\%$&12.29\\\cline{2-6}
		&10.831&0.080&3&96.15$\%$&17.62\\\cline{2-6}
		&10.831&0.080&5&96.15$\%$&28.17\\\hline
	\end{tabular} 
\end{table*} 
    We processed 13 sets of holding-the-hand action and 16 protest-the-music action, and corresponding obtained 130 groups of holding-the-hand samples and 160 groups of protest-the-music samples. Select 70 groups and 90 groups respectively from these two kinds of samples as training sets, and the remaining 60 groups and 70 groups as test sets. The penalty factor $c$ and kernel function parameter $\sigma$ of $svm$ model are determined by grid search and quantum genetic algorithm, respectively. In this paper, we won't use any dimensional reduction algorithm. It mainly bases on two reasons: Firstly, this is for better expansibility. It can extend from the upper limb action classification to the whole body action classification. Secondly, in our classification algorithm, the time and the computing complexity is acceptable for $ 22 $-dimensional data. Thus, we take the sample directly into the SVM for training. Following, The holding-the-hand action will be labelled as $ 1 $, the protest-the-music action will be labelled as $ 2 $.
    
    During the calculation process, we set the population size to 80 and the quantum bit length to 60 for QGA. The search range of penalty factor $ C $ is set to $[2^{-2},2^{4}]$, meanwhile the range of the kernel function parameter $\sigma$ is set to $[2^{-4},2^{4}]$. We also provide the classification accuracy results for different generations of quantum genetic algorithm. The results are shown in Table~\ref{result}. It can be seen that quantum genetic algorithm almost converges after two generations. For this reason, we won't consider catastrophe situation here, and set parameter $\epsilon=0$. Due to the fast convergence speed, we can see that the time complexity difference between grid research and QGA is not very large. Otherwise, It's noted that the quantum genetic algorithm increased the classification accuracy by nearly $ 2.5 \% $ at the expense of less time. As we know, quantum algorithm has the parallel characteristic, it can search much more larger parameter space with the same time. So the more optimized $C$ and $\sigma$ can be found with the help of quantum method improvement. For the grid research approach, a very large amount of computation will be needed to achieve such an accuracy. To make the results more intuitive, we refine the results by confusing matrices in Table \ref{confusion_matrix_1} and Table \ref{confusion_matrix_2}.
    
    \begin{table}[ht]
    	\centering    
    	\caption{Confusion Matrix}
    	\label{confusion}
    	\subtable[QGA]{   
    		\begin{tabular}{|c|c|c|c|}\hline  
    			\diagbox{}{}& Throw & Raise both arms&Accuracy\\\hline
    			Throw &60&0&$100\%$\\\hline
    			Raise both arms &5&65&$92.86\%$\\\hline    
    		\end{tabular}    
    		\label{confusion_matrix_1}
    	}
    	\qquad
    	\subtable[CV]{        
    		\begin{tabular}{|c|c|c|c|}\hline 
    			\diagbox{}{}&Throw & Raise both arms&Accuracy\\\hline
    			Throw &60&0&$100\%$\\\hline
    			Raise both arms &8&62&$88.57\%$\\\hline
    		\end{tabular}
    		\label{confusion_matrix_2}
    	}
    \end{table}
    
The confusion matrix is shown in Table~\ref{confusion}. The solution space of grid search is limited and the result is farther from the optimal solution. The quantum genetic algorithm takes the characteristics of quantum parallelism, extends the solution space at the cost of a little higher time complexity and brings better results. On the other hand, we can use the angles of limbs attached to the joints to represent and identify the human behavior pattern and the correct rate of this method can achieve an accuracy of above $95\%$. On some conditions, this can be thought as a successful classification result.

\section{Conclusions}
    
With the help of the parallel characteristics of quantum algorithm, we succeeded in improving the accuracy of SVM classification at cost of a little time complexity. This paper can be considered as a good example of the combination of QGA and classification algorithm. The quantum-inspired algorithm can also be used in combination with other algorithms. Next, we will work on more complex actions and search new features to further improve the accuracy of SVM classification.
    
This paper presents a new method of representing and classifying human actions by using the quantum generic algorithm to optimize the parameter of the SVM. We extracted the joints' angles from the skeleton joints' positions to represent the human stick figure in Kinect. By this way, the dimensionality was reduced by $ 1/3 $. By reducing the dimensionality of samples and increasing the efficiency of computation, we achieved a higher classification accuracy in comparison with the conventional pattern recognition method.

\section{Acknowledgement}
This research has been funded by the National Key Research and Development Plan under the grant 2017YFC0804401; the Natural Science Funds of Jiangsu Province of China under Grant BK20140216; and the National Key Research and Development Plan under the grant 2017YFC0804401.

\bibliography{refs}

\end{document}